\documentclass[onecolumn]{IEEEtran}

\usepackage{xcolor}

\usepackage{amsopn}

\usepackage{lmodern}
\usepackage{amsmath}

\usepackage{graphicx}

\usepackage{amssymb,amsmath}

\usepackage{relsize}

\usepackage{upgreek}

\usepackage{subfigure}

\usepackage{url}

\usepackage{graphicx}

\usepackage{multicol}

\usepackage{algorithmic}
\usepackage{algorithm2e}

\pdfoutput=1
\usepackage{amsmath, amssymb}
\usepackage{graphicx,psfrag,epsf}
\usepackage{enumerate}
\usepackage[square,numbers]{natbib}
\usepackage{url} % not crucial - just used below for the URL 

\newcommand{\argmax}{\mathop{\rm argmax}}

\def\X{\mathcal{X}}

\def\R{\mathbb{R}}

\def\x{\mathbf{x}}

\def\u{\mathbf{u}}

\def\U{\mathbf{U}}

\def\u{\mathbf{u}}

\def\Y{\mathcal{Y}}

\def\R{\mathbb{R}}
\def\x{\mathbf{x}}
\def\knn{$k$NN}
\def\X{\mathcal{X}}

\def\U{\mathbf{U}}
\def\u{\mathbf{u}}

\newtheorem{remark}{Remark}

\ifCLASSINFOpdf
\else
\fi

\usepackage{multicol}
\usepackage{lipsum}

\newcommand\blfootnote[1]{%
  \begingroup
  \renewcommand\thefootnote{}\footnote{#1}%
  \addtocounter{footnote}{-1}%
  \endgroup
}

\begin{document}
\title{Clustering by Non-parametric Smoothing}

\author{
David~P.~Hofmeyr
        % <-this % stops a space
\thanks{D.P. Hofmeyr was with the School of Mathematical Sciences, Lancaster University, United Kingdom.\\
\email{d.p.hofmeyr@lancaster.ac.uk}\\
\\
 \textit{\textcopyright 2025 IEEE. Personal use of this material is permitted. Permission from IEEE must be obtained for all other uses, in any current or future media, including reprinting/republishing this material for advertising or promotional purposes, creating new collective works, for resale or redistribution to servers or lists, or reuse of any copyrighted component of this work in other works.}}% <-this % stops a space
% <-this % stops a space
}

\begin{center}
{\Huge \bf Bags of Projected Nearest Neighbours: Competitors to Random Forests?}
\end{center}
\vspace{20pt}
{\bf David P. Hofmeyr\\
 School of Mathematical Sciences, Lancaster University, UK}
\vspace{20pt}

\begin{multicols}{2}

\begin{abstract}
In this paper we introduce a simple and intuitive adaptive $k$ nearest neighbours classifier, and explore its utility within the context of bootstrap aggregating (``bagging''). The approach is based on finding discriminant subspaces which are computationally efficient to compute, and are motivated by enhancing the discrimination of classes through nearest neighbour classifiers. This adaptiveness promotes diversity of the individual classifiers fit across different bootstrap samples, and so further leverages the variance reducing effect of bagging. Extensive experimental results are presented documenting the strong performance of the proposed approach in comparison with Random Forest classifiers, as well as other nearest neighbours based ensembles from the literature, plus other relevant benchmarks. Code to implement the proposed approach is available in the form of an {\tt R} package from \url{https://github.com/DavidHofmeyr/BOPNN}.
\end{abstract}

\begin{IEEEkeywords}
Classification; adaptive \knn; nearest neighbours; ensemble model; discriminant subspace; bootstrap aggregating; bagging.
\end{IEEEkeywords}

\blfootnote{\textit{\textcopyright 2025 IEEE. Personal use of this material is permitted. Permission from IEEE must be obtained for all other uses, in any current or future media, including reprinting/republishing this material for advertising or promotional purposes, creating new collective works, for resale or redistribution to servers or lists, or reuse of any copyrighted component of this work in other works.}}

% \fi
\IEEEpeerreviewmaketitle

\section{Introduction}
\label{sec:intro}

Bootstrap aggregating~\citep{breiman1996bagging}, or ``bagging'', is the approach of combining the outputs of several predictive models, each fit to different bootstrap samples from a set of data, into a single \textit{ensemble} predictive model. Bagging has remarkable potential for improving the prediction performance of high variance predictors, due to the variance reducing effect of model averaging. However, it is well understood that not all high variance predictors are able to leverage this effect equally, due to some being ``too stable'' across different bootstrap samples. 
%
%For example, when using the standard \knn~model within a bagged ensemble, in order to achieve optimal convergence rates the proportion of the sample to be included in each bootstrap sample must tend to zero in order. 
%
%Indeed it is known that for the  The predictions from such models have high pairwise correlation, meaning that averaging their predictions does little to reduce the overall variance.
%
Bagged ensembles of Decision Tree (DT) based models have undeniably shown the greatest promise to date, to the extent that bagging is sometimes categorised as a decision tree based approach~\citep{ESL}. The remarkable success of Random Forest~\citep[RF]{breimanRF} based models has only further entrenched DTs as the \textit{de facto} ``optimally baggable'' model. RFs differ from regular DTs only through the addition of a randomisation step preceding each stage in the standard Classification And Regression Trees~\citep[CART]{breiman2017classification} algorithm. However, this simple modification has a remarkable ``destabilising'' effect on the already highly variable trees, and so enables further variance reduction through averaging.\\
\\
Attempts have been made to emulate the success of RFs and bagged DTs with other non-parametric models, such as those based on nearest neighbours~\citep{zhou2005adapt, RPEnsemble, esknn}. Although these have shown some success, it is questionable whether any of these approaches has the potential to be a real competitor to RFs across many different settings, partly due to limited experimental results having been documented. %In this paper we investigate this topic in greater depth.

%A number of attempts have been made to produce ensembles similar to bagged models, combining other flexible non-parametric models, such as those based on nearest neighbours~\citep{zhou2005adapt, RPEnsemble, esknn}. Although these have shown some success, it is questionable whether any of these approaches has the potential to be a real competitor to RFs across many different settings, partly due to limited experimental results having been documented. In this paper we investigate this topic in greater depth.

It is well known that a necessary condition for the success of a bagged ensemble is substantial diversity in the models~\citep{krogh1994neural}, however existing approaches for inducing this diversity in \knn~ensembles have largely been based on randomisation alone~\citep{zhou2005adapt, domeniconi2004nearest, deegalla2022random}, possibly owing to the successful application of randomisation within RFs. But purely randomised methods can only be beneficial if the resulting increase in diversity across models substantially outweighs the decrease in accuracy of the individual models caused by the extra randomness. Where RFs are fundamentally advantaged over these nearest neighbours based ensembles, however, is in the adaptive way in which DTs determine their ``smoothing neighbourhoods'', with the additional randomisation being a secondary factor. The adaptiveness of DTs simultaneously is a primary source of diversity in the models beyond the randomness of the bootstrap sampling, and also reduces the amount by which the added randomness of RFs impacts on the quality of the individual models.\\
\\
In this paper we introduce an intuitive adaptive $k$~Nearest~Neighbours~(\knn) classifier which is computationally efficient to compute, and explore its utility within bagged ensembles. %The approach which we propose is relatively simple, and we recognise the potential for improvement.
Our main objectives with this piece of work are to illustrate the importance of, and potential offered by, including an adaptive learning step within each model in a bagged \knn~ensemble; and to support this illustration with a rigorous and extensive set of experiments.\\
\\
The remainder of this paper is organised as follows. In the next section we provide some background on bagging, with particular focus on its application to \knn~based models. In Section~\ref{sec:method} we describe our approach, as well as some of the practicalities surrounding implementation and useful outputs from the resulting models. In Section~\ref{sec:experiments} we document the results from experiments using all 162 data sets in the Penn Machine Learning Benchmarks repository~\citep[PMLB]{PMLB}, in which we compare the performance of the proposed approach with RFs, as well as numerous other models for context. Random Forest classifiers are viewed by many as excellent general purpose models; seldom much worse than any others, and frequently among the best performing models on data from extremely diverse domains. We are of the firm opinion that in order to support any new model as a realistic alternative in this regard, there should be no possibility of data set selection bias (whether conscious or unconscious) which is possible whenever any subset of available data sets without a clear and justifiable selection criterion is used.

%\\
%\\
%In the next section we give additional details related to the principles of bagging and \knn~based models, as well as describing some existing attempts to combine these into effective predictive models. In Section~\ref{sec:method} we provide details on the proposed method, as well as some discussion regarding its formulation. In Section~\ref{sec:experiments} we provide a detailed analysis of the results of an extensive set of experiments across a very large collection of (162) publicly available data sets. This is the entire collection of classification data sets from the Penn Machine Learning Benchmarks repository~\citep[PMLB]{PMLB}. 

\section{Bagging, \knn, and What's Been Tried} \label{sec:background}

In this section we provide light technical background on bootstrap aggregating, and discuss some of its applications to \knn~based classifiers.
Suppose we receive a sample of observations, say $D:= \{(\x_1, y_1), (\x_2, y_2), ..., (\x_n, y_n)\}$, assumed to have arisen independently from some probability distribution function, $F_{X,Y}$, on $\R^d\times [K]$, where we have used $[K]$ to denote the first $K$ natural numbers, i.e., $[K]:= \{1, ..., K\}$. That is, the \textit{response variables} (or ``class labels''), $y_i; i \in [n]$, each takes on one of $K$ known and distinct values, and the associated observations of the \textit{covariates}, $\x_i; i \in [n]$, are each $d$-dimensional real vectors. The classification task is then to use this \textit{training} sample, $D$, to obtain a model $g(\cdot |D)$ which, given a \textit{query point}, $\x \in \R^d$, is able to provide a prediction for the class to which $\x$ belongs.

Bagging operates by resampling from $D$ multiple times to produce $B$ \textit{bootstrap} samples, $\tilde D_1, ..., \tilde D_B$, and then combining the resulting models, $g(\cdot |\tilde D_b); b = 1, ..., B$, to obtain a final predictive model. Note that whether each bootstrap sample is obtained by sampling with or without replacement often has relatively little impact on the performance of the overall model, and for ease of exposition we consider sampling without replacement, and in such a way that each bootstrap sample contains $n_B < n$ observations.

%and the task is to use this \textit{training} sample to obtain a model, say $g(\cdot|D)$, which, given a \textit{query point} $\x \in \X$ is able to provide a prediction for a corresponding value for $y \in \Y$. We have used the notation $g(\cdot|D)$ to capture the explicit dependence of the model on the training data, $D$, in that we may imagine the object $g(\cdot|\cdot)$ as an algorithm which, upon receipt of a sample, outputs a function on $\X$. As in the case of Random Forests, we also allow for some randomness in how this algorithm operates.

\subsection{Bagging and Variance Reduction}

Although requiring considerably greater computational investment than fitting a single predictive model, bagging can be remarkably effective in reducing the variance of flexible predictive models. This is most conveniently communicated when combining the individual models through averaging, i.e., when using
\begin{align}\label{eq:bag_average}
    g^{(bag)}(\x|D) = \frac{1}{B}\sum_{b=1}^B g(\x|\tilde D_b).
\end{align}
\begin{remark}
    Although averaging classification models whose outputs are themselves class labels is clearly nonsensical, the averaging formulation in Eq.~(\ref{eq:bag_average}) is applicable in this context if we simply consider that each model either outputs an indicator vector for the class allocation of an input vector, $\x$, or an estimate for the conditional distribution of $Y|X=\x$. In the latter case the quantity $g^{(bag)}(\x|D)$ may also be seen as an estimate for the distribution of $Y|X=\x$, whereas in the former $g^{(bag)}(\x|D)$ is better interpreted as an estimate for the distribution of $g(\x|\tilde D_0)$, where $\tilde D_0$ is a sample of size $n_B$ drawn directly from the underlying population. In either case a final class prediction can be obtained by taking the mode of $g^{(bag)}(\x|D)$.
\end{remark}

\noindent
Now, it is straightforward to show that~\citep{ESL},
\begin{align}\label{eq:bag_benefit}
    Var(g^{(bag)}(\x|D)) = Var(g(\x|\tilde D_1))\left(\rho + \frac{1-\rho}{B}\right),
\end{align}
where $\rho = Cor(g(\x|\tilde D_1), g(\x|\tilde D_2))$. Note that although each $\tilde D_b; b\in[B]$, has the same distribution as an i.i.d.\footnote{recall that we consider resampling without replacement} sample of the same size drawn directly from the underlying population, it is not the case that the joint distributions of any pair $\tilde D_{b_1}, \tilde D_{b_2}; b_1\not = b_2$ are the same as those of pairs of independent samples from the population due to the (potential) overlap of the bootstrap samples. It is this fact which results in $Cor(g(\x|\tilde D_1), g(\x|\tilde D_2))$ generally being greater than zero. Where bagging is typically at its most beneficial, then, is when the bias and variance of $g(\x|\tilde D_1)$ are similar to those of $g(\x|D)$, i.e., using a smaller sample does not affect accuracy too substantially, \textit{and} where $Cor(g(\x|\tilde D_1), g(\x|\tilde D_2))$ is relatively small. It is also important to recognise that bagging does not improve on the bias of the model, and so bagging inflexible models which have low variance but may have high bias does not provide much benefit to the overall prediction accuracy.

%For convenience we reserve the possibility that the output of the model, $g(\x|D)$, is not itself an explicit prediction for the class to which $\x$ belongs, but rather can be used very easily to obtain such a classification. For example, we allow for $g(\x|D)$ to represent an estimate for the entire conditional distribution of $Y|X=\x$, or a $C$-length vector of zeroes except in the position of the desired prediction for $y$, where it takes the value one. For the lattes, if $C = 4$ and our prediction is that $\x$ is associated with class $3$, we may have $g(\x|D) = (0, 0, 1, 0)$. This is a convenience which allows us to express the bagged classification model as an average of multiple such models, as described in the following.

Where bagging really shines is when applied to flexible, low-bias models, between which the correlation due to overlapping samples is relatively low. Generally speaking the class of non-parametric smoothing models can be made extremely flexible by selecting a small ``smoothing parameter''. For example, the \knn~model bases its prediction for a point, $\x$, only on the properties of the nearest $k$ points to $\x$ from among the $\x_i; i \in [n]$. However, \knn, and other so-called ``lazy learners'', have been referred to as ``too stable'' from the point of view of bagging, because the correlation induced by overlapping bootstrap samples is substantial. This can be intuited by considering the \textit{region of influence} of an observation, say $\x_i$, as the subset of $\R^d$ to which $\x_i$ is one of the $k$ nearest from among the sample. Note that this region is completely independent of the observations of the response variable, and may depend on only a very small number of other sample points. This independence of the responses means the standard \knn~model is not able to leverage the relationships between the covariates and the response in order to improve its fit (hence the term ``lazy learner''). The extreme localisation implied by the fact that the region of influence of a point depends on so few other points is also why the \knn~predictions from two samples with substantial overlap are so correlated. Indeed it has been shown that in order to decrease the correlation sufficiently to achieve optimal convergence rates, 
the (expected) proportion of overlap between pairs of bootstrap samples must tend to zero, which can only be achieved if the proportion of the sample included in each bootstrap sample itself tends to zero~\citep{samworthBNN}. Decision trees, on the other hand, are \textit{adaptive} non-parametric smoothers, and aggressively exploit the relationships between the covariates and response in how they recursively split up the input space to actively determine the regions of influence of each point. In this way the region of influence of each point can be dependent on every other point in the sample and, as a result, the \textit{non-overlapping parts} of two bootstrap samples are able to differentiate their respective models sufficiently to induce lower correlation between their predictions.

\subsection{Bagged \knn~Classifiers}

As mentioned previously, standard \knn~models are seen to be stable from the point of view of bagging. However, there is a prevailing opinion that they can be ``destabilised'' by adding randomisation to the way in which distances are determined within the nearest neighbour search for each model. This can be achieved in a number of ways, such as only computing distances on a (random) subset of the variables in $\R^d$~\citep{domeniconi2004nearest}; by randomly projecting the observations before computing distances~\citep{deegalla2022random}; or by using a random selection of the value of $p$ within the $L_p$-norm derived distance function itself~\citep{zhou2005adapt}. By modifying the distance metric, a greater variety of points can have an impact on the regions of influence of others. However, none of these approaches is adaptive to the relationships between the covariates and the response, and there is insufficient evidence that purely randomised approaches, such as these, are useful in general. Ultimately, since the dominant term in $Var(g^{(bag)}(\x|D))$ is equal to $\rho Var(g(\x|\tilde D_1))$, a modification such as this can only be beneficial if the reduction in $\rho$ outweighs the increase in the variance of the individual models, which may be substantial if the modification is purely random.

Nearest neighbours models can be made adaptive by 
%, for example, selecting the value of $k$, either globally or separately by location, according to some validation strategy~\citep{variable_knn}; or by
actively learning a distance metric to enhance discrimination of classes, either globally~\citep{NCA} or locally~\citep{hastie1995discriminant}. As far as we are aware, however, no such approaches have been explored within the context of bagging, likely because of the computational demand of fitting a large number of such models. %Generally speaking these methods are considerably more computationally demanding than the standard \knn~model, and so their use within bagged ensembles is fairly restricted. 
Somewhere between fully adaptive and randomised is the approach of selecting, based on some measure of predictive ability, from among multiple \knn~models arising from different random projections of the observations. This approach has shown success in the context of bagging~\citep{RPEnsemble}, however the number of individual \knn~models is equal to the product of the number of bootstrap samples and the number of random projections from among which to select each model in the ensemble, which is a considerable computational restriction. A related method, which also uses a selection from multiple randomised \knn~models, first fits a very large number of models to different bootstrap samples with each using its own random selection of the variables in $\R^d$ for computing distances, and then selects a fixed proportion of these for inclusion in the ensemble~\citep{esknn}. Very importantly, since the selection of each model in the final ensemble is from the same collection of candidates, the models in the ensemble are strongly dependent on one another. To counteract this, the selection of models in the ensemble is not purely based on their apparent predictive ability, but is also made to ensure some level of differentiation in the predictions across the different models. Although the total number of models to be fit is substantially fewer than when a fully independent selection is made for each model in the ensemble, this approach is still substantially slower than alternatives. Moreover, this approach loses the statistical ``niceness'' of bagging, and in particular does not provide any Out-Of-Bag (OOB) estimates for performance. This further limits the applicability of this approach when any substantial hyperparameter search is needed to obtain a good model.

%Moreover, as documented in Section~\ref{sec:results}, we have found the performance of this approach to be substantially inferior to relevant competitors.

\section{Bagging Adaptive \knn~Classifiers Based on Discriminant Projections}\label{sec:method}

As discussed previously, the fundamental advantage which DTs have over other non-parametric smoothing models, within the context of bagging, is the adaptive determination of the regions of influence of each point. This benefit manifests both through how it allows DTs to exploit the relationships between the covariates and the response, and also through the fact that the region of influence of a point is dependent on the entire sample. This latter fact induces further differentiation across the outputs of DTs fit to different bootstrap samples, and so reduces their correlation. 

Making \knn~classifiers adaptive by optimising the distance metric used in the nearest neighbour search is intuitively pleasing, and in principle has the potential to achieve the same benefits as those described above for DTs. But where these adaptive \knn~methods are limited is their computation time, which pales in comparison with the speed of DTs, and all but precludes their use within bagged ensembles. A fact which seems largely to have been overlooked, however, is that adaptively modifying the distance metric to enhance class discrimination does not have to be performed in a fully optimal manner in order to leverage the benefits mentioned above. 
We therefore explore this potential through a more computationally efficient alternative, based on finding discriminant subspaces designed to enhance class discrimination as determined by \knn~classifiers. This is similar to metric learning in that computing Euclidean distances within a chosen $q$-dimensional subspace, say with basis vectors in the columns of a matrix $\U \in \R^{d\times q}$, is equivalent to computing the Mahalanobis distance with weight matrix $\U\U^\top$, defined as $d(\x, \x') = \sqrt{(\x-\x')^\top \U\U^\top (\x - \x')}$.

%has frequently been approached through ``metric learning''~\citep{adaptiveKnn}, in which the distance function used for the nearest neighbour search is optimised for the classification task. Most often this is achieved through optimising the scaling matrix, $S$, within the Mahalanobis distance defined by $d_S(\x, \x^\prime) := \sqrt{(\x-\x^\prime)^\top S(\x-\x^\prime)}$. When this metric is applied globally, the adaptiveness has the potential to achieve the same benefits as those described above, which DTs enjoy, for effective use within bagging. However, where these existing adaptive \knn~methods are limited is their computational complexity, which pales in comparison with the speed of DTs. In this section we introduce a simple and intuitive adaptive \knn~classifier which is based on finding a discriminant subspace which is motivated by enhancing the accuracy of the \knn~classifier, and is computationally efficient to compute.

\subsection{A Simple Discriminant Subspace for \knn}

Discriminant subspaces are subspaces of $\R^d$ within which the classes are (relatively) easily separated from one another. Depending on the classifier being applied after projection, different formulations of the discriminant subspace will be more/less appropriate. For example, the well known Linear Discriminant Analysis~\citep[LDA]{LDA} model, in which each class is modelled with a Gaussian distribution and all classes share a common covariance matrix, has a so-called ``sufficient subspace'' given by the eigenvectors of $\hat \Sigma_w^{-1}\hat \Sigma_b$ associated with its non-zero eigenvalues. Here $\hat \Sigma_w$ is the pooled within-class covariance estimate and $\hat \Sigma_b = \hat \Sigma - \hat \Sigma_w$, where $\hat \Sigma$ is the overall data covariance. A more flexible alternative, known as Mixture Discriminant Analysis~\citep[MDA]{MDA}, models each class with a Gaussian mixture. If all mixture components across all classes share a common covariance matrix, then a similar discriminant subspace can be obtained. When more general formulations are adopted, discriminant subspaces can be obtained by maximising the \textit{classification likelihood}, given by the multinomial likelihood with probabilities determined using Bayes rule and with the class densities estimated on the projected data~\citep{peltonen2005discriminative,hofmeyr2024optimal}. These classification likelihood based approaches require numerical optimisation, and are thus not computationally competitive with those which can be obtained using highly optimised eigen-solvers.

Generally speaking, however, discriminant subspaces can be thought of as pushing points into high density regions within their own classes, and into low density regions within other classes; and the appropriateness of a subspace depends on how density is being modelled.
Motivated by this simple but principled idea, we adopt the following heuristic, which has some similarity with a fully non-parametric MDA. For each $i$, we let $i_k$ be the $k$-th nearest observation to $\x_i$ from within its own class, and $i'_k$ the $k$-th nearest observation to $\x_i$ from among those in other classes. We then define
\begin{align}
    \hat \Sigma_{in} &:= \frac{1}{n}\sum_{i=1}^n (\x_i - \x_{i_k})(\x_i - \x_{i_k})^\top,\\
    \hat \Sigma_{out} &:= \frac{1}{n}\sum_{i=1}^n (\x_i - \x_{i'_k})(\x_i - \x_{i'_k})^\top.
\end{align}
%$\hat \Sigma_{in} = \frac{1}{n}\sum_{i=1}^n (\x_i - \x_{i_k})(\x_i - \x_{i_k})^\top$ and $\hat \Sigma_{out} = \frac{1}{n}\sum_{i=1}^n (\x_i - \x_{i'_k})(\x_i - \x_{i'_k})^\top$.
For a unit vector $\u \in \R^p; ||\u|| = 1$ the quantity $\u^\top \hat \Sigma_{in}\u$ (respectively $\u^\top \hat \Sigma_{out}\u$) is then the average squared distance from each point to its $k$-th \textit{same class neighbour} (respectively \textit{other class neighbour}), measured along direction $\u$. Such unit vectors which lead to small values of $\u^\top \hat \Sigma_{in}\u$ and large values of $\u^\top \hat \Sigma_{out}\u$ are thus desirable \textit{discriminant directions} for a \knn~classifier. A sensible discriminant subspace is therefore formed by simply taking the leading eigenvectors of $\hat \Sigma_{in}^{-1}\hat \Sigma_{out}$.

\begin{remark}
    Although the quantities $\u^\top \x_{i_k}$ and $\u^\top \x_{i^\prime_k}$ will tend not to be precisely the $k$-th nearest in- and out-of- class neighbours to $\u^\top \x_i$ (i.e., the ordering of distances changes after projection onto $\u$), they nonetheless tend to be from among the nearer in- and out-of-class points, and so minimising the post-projection in-class and maximising the post-projection out-of-class distances still has the desired effect.
\end{remark}

\begin{remark}
    The quantity $\hat \Sigma_{in}$ can be seen as capturing the average \textit{local} within class covariance, and is similar to the average within component covariance matrix used in MDA with a very large number of components. The analogous subspace arises from the eigenvectors of $\hat \Sigma_{in}^{-1}(\hat \Sigma - \hat \Sigma_{in})$, however we have found the formulation described above to yield better performance within bagged ensembles. Intuitively the leading eigenvectors of $\hat \Sigma_{in}^{-1}(\hat \Sigma - \hat \Sigma_{in})$ result in projections on which the in-class neighbours are pushed nearer to one another, due to the term $\hat \Sigma_{in}^{-1}$, while the term $(\hat \Sigma - \hat \Sigma_{in})$ ensures this doesn't arise simply by pushing {\em all} points closer to one another. On the other hand the term $\hat \Sigma_{out}$ in our approach explicitly contributes to class discrimination by pushing points away from their out-of-class neighbours. Note that the matrix $\hat \Sigma_{in}^{-1}\hat \Sigma_{out}$ also uses a greater amount of information from the sample than does $\hat \Sigma_{in}^{-1}(\hat \Sigma - \hat \Sigma_{in})$, and therefore intuitively leads to greater diversity across bootstrap samples.
\end{remark}

\subsection{Other Practicalities}

In this subsection we comment on some further practical aspects of the proposed approach of bagging the predictions arising from \knn~models obtained on different discriminant subspaces. We use, for each $b \in [B]$, the notation $\hat \Delta_{b}$ to represent the \textit{discriminant matrix} whose leading eigenvectors, in the columns of the matrix $\U_b$, determine the discriminant subspace for the $b$-th model in the ensemble. The matrix $\hat\Delta_b$ represents the matrix $\hat \Sigma_{in}^{-1}\hat \Sigma_{out}$ computed from the $b$-th bootstrap sample, however we introduce this new notation to accommodate a slight deviation which we describe below.

\subsubsection{Additional Diversity Through Randomisation}

As described previously, additional randomness across different models (e.g. through randomised variable selection) can be beneficial if the resulting decrease in accuracy of the models arising from this randomness (which may also be seen as a reduction in information) is outweighed by the effect of additional diversification of the individual models. Adaptive models mitigate the reduction in accuracy arising from reduced information, when compared with lazy learners, as they better exploit what information \textit{is} available and are better suited to filtering out noise.

We have found that within our proposed approach restricting the discriminant subspaces each to lie within their own randomly determined higher dimensional subspace does indeed improve performance in general. How we implement this is to, for each $b \in [B]$, randomly sample a subset of the covariates, $Q_b \subset [d]$, of size $q_0 \leq d$, and then compute $\hat \Sigma_{in, b}$ and $\hat \Sigma_{out, b}$ from the $b$-th bootstrap sample but restricted to the subspace containing the variables in $Q_b$. The discriminant matrix $\hat \Delta_b$ is then simply equal to $\mathbf{I}_{Q_b}\hat \Sigma_{in, b}^{-1}\hat \Sigma_{out,b}\mathbf{I}_{Q_b}^\top$ where $\mathbf{I}_{Q_b} \in \R^{d \times q_0}$ has as columns the cardinal basis vectors for the variables in $Q_b$.

This randomised reduction in dimensionality is beneficial both from the point of view of classification accuracy of the overall ensemble, and importantly also substantially reduces the computational complexity of the method. This is because nearest neighbour search and matrix inversion only needs to be performed in dimension at most $q_0$, and for large $d$ we have found $q_0 \propto \sqrt{d}$ is more than adequate to achieve high accuracy.

\subsubsection{Variable Importance}

Interpretability of flexible predictive models is increasingly a point of focus, as modern methods typically rely on intricate relationships between the covariates and the response variable which may not be explicitly expressed within the model in any intelligible form. Variable importance scores are measures of the overall contribution of the covariates to the predictions made by a model. Although far from encapsulating the entirety of what a model has captured in the data these are nonetheless useful diagnostics for understanding which are the important variables driving the model's predictions.\\
\\
The discriminant subspace framework offers an intuitive means by which the contribution of each variable to the model predictions may be quantified. In particular, if a variable lies within the discriminant subspace from one of the models in the ensemble, then it is natural to view that variable as important to the predictions from that particular model. On the other hand if a variable lies in the orthogonal complement of the subspace then this variable is unimportant. In general none of the variables will lie entirely within/without any of the discriminant spaces, but there will be some non-zero angles between them. If $\U$ has as columns a normalised (not necessarily orthogonal) basis for a subspace, then the cosines of the principal angles between the subspace and each of the variables in the cardinal basis can be seen as capturing the importance of each variable to the subspace and lie in the diagonal elements of $\U\U^\top$. To further capture the relative discriminatory information in the dimensions of the subspace, we weigh the basis vectors, which arise from the spectral decomposition of the discriminant matrix, by the eigen\textit{values}. To determine the importance of the $j$-th variable to the entire ensemble we then average its importance values from each of the subspaces in the model. Specifically, if $\hat \Delta_b = \U_b\boldsymbol{\Lambda}_b\U_b^{-1}$ is the standard spectral decomposition of the $b$-th discriminant matrix, with eigenvalues in the diagonal matrix $\boldsymbol{\Lambda}_b$ listed in decreasing order, then the importance of the $j$-th variable to the predictions from the ensemble are given by $\frac{1}{B}\sum_{b=1}^B (\mathbf{V}_b\mathbf{V}_b^\top)_{jj}$, where $\mathbf{V}_b \in \R^{d\times q}$ has as columns the first $q$ columns of $\U_b\boldsymbol{\Lambda}_b^{1/2}$, and where $q \leq q_0$ is the dimension of the discriminant subspaces.

\begin{figure*}%[t]
    \centering
    \subfigure[Subspace from a single model using the proposed approach]{\includegraphics[height = .15\linewidth, width=0.24\linewidth]{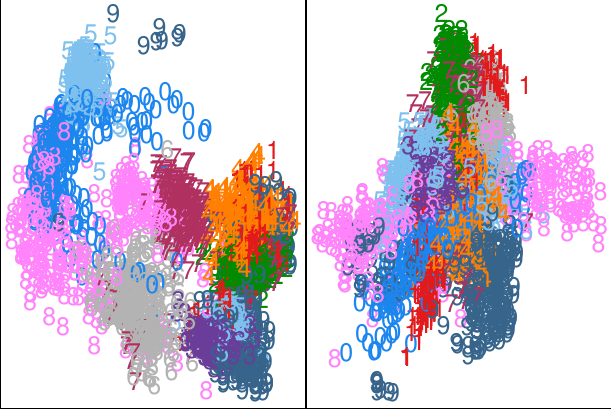}\includegraphics[height = .15\linewidth, width=0.24\linewidth]{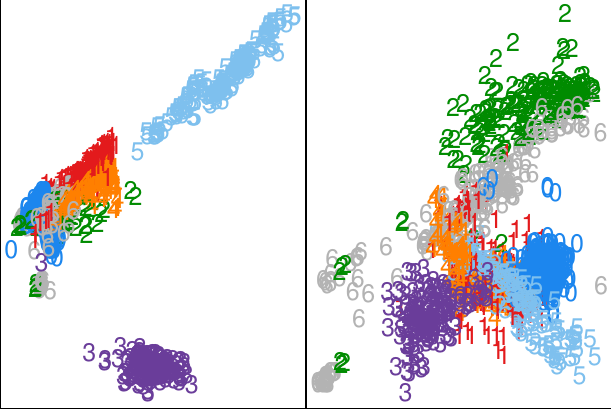}\label{fig:projections1}}
    \subfigure[Subspace from the proposed ensemble, using the PCA basis from the averaged projections]{\includegraphics[height = .15\linewidth, width=0.24\linewidth]{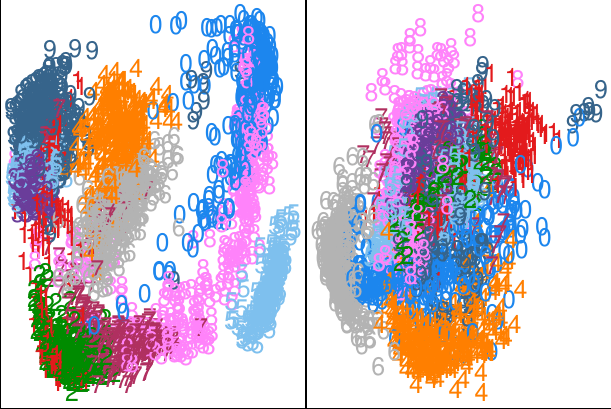}\includegraphics[height = .15\linewidth, width=0.24\linewidth]{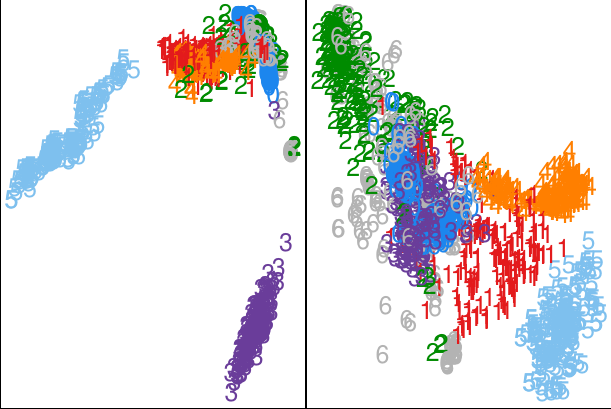}\label{fig:projections2}}
    \subfigure[Linear Discriminant Analysis subspace]{\includegraphics[height = .15\linewidth, width=0.24\linewidth]{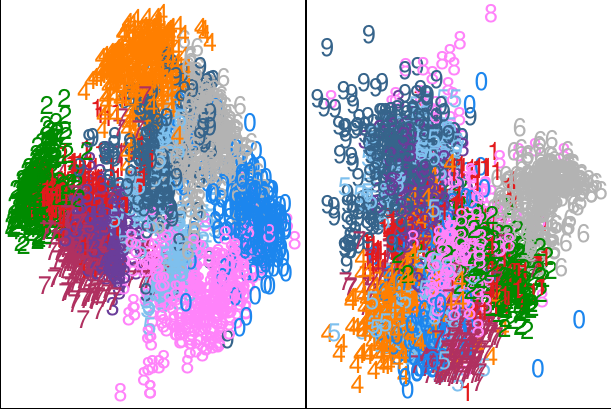}\includegraphics[height = .15\linewidth, width=0.24\linewidth]{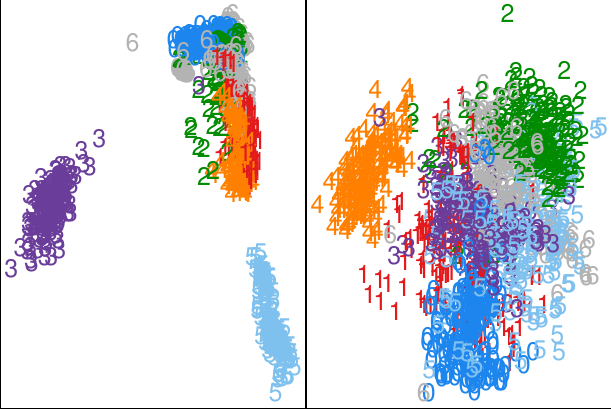}\label{fig:projections3}}
    \subfigure[Mixture Discriminant Analysis subspace]{\includegraphics[height = .15\linewidth, width=0.24\linewidth]{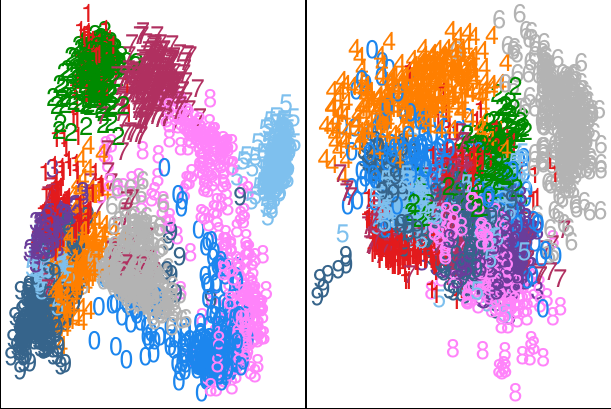}\includegraphics[height = .15\linewidth, width=0.24\linewidth]{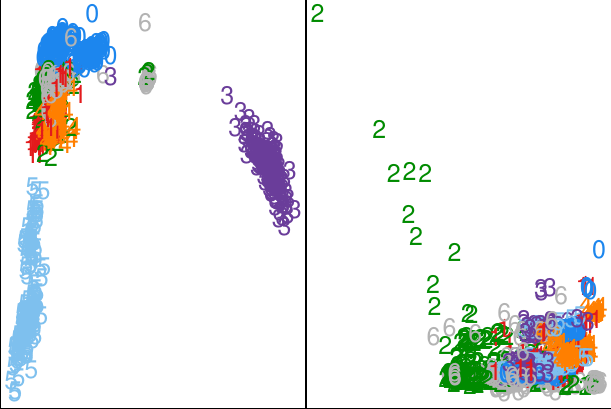}\label{fig:projections4}}
    %\subfigure[Accuracy]{\includegraphics[width=0.48\linewidth]{acc.pdf}}
    %\subfigure[FDA (0.921, 0.985)]{\includegraphics[width=0.48\linewidth]{fda.pdf}}
    \caption{2-Dimensional projections of the pen-based recognition of handwritten digits and image segmentation data sets.}
    \label{fig:projections}
\end{figure*}

%\begin{multicols}{2}

\subsubsection{Visualisation}

A further advantage of discriminant subspaces is the fact that they reduce the dimensionality of the observations, which can be beneficial for obtaining visualisations of the classes, and their separations from others, as well as the predictions made by a model. A single discriminant matrix, $\hat \Delta_b; b \in [B]$, is subject to fairly high variation, and moreover how to select from among multiple discriminant subspaces, within the ensemble model, to obtain a single visualisation is not immediately obvious. We therefore aggregate the information from all discriminant subspaces, by determining the ensemble projection $\bar{\mathbf{P}}:= \frac{1}{B} \sum_{b=1}^B\U_{b,1:q}\U_{b,1:q}^\top$, where $\U_{b,1:q}$ is the first $q$ columns of $\U_b$. To obtain a visualisation of the observations in the aggregated discriminant subspace we project them onto the principal components basis vectors computed from the aggregate projected observations $\{\tilde \x_i\}_{i\in [n]}; \tilde \x_i := \bar{\mathbf{P}}\x_i$.

Figure~\ref{fig:projections} shows two examples, where projections of two of the data sets used in our experiments are shown. Each sub-figure contains four plots, with the left pair showing the first four discriminant projections of the sixteen dimensional pen-based recognition of handwritten digits data set~\citep{pendigits}, and the right pair those of the nineteen dimensional segmentation data set~\citep{imageseg}. The points depict the projections of an independent test set separate from the ``training'' set used to obtain the actual projection directions, and the colours and point characters represent the individual classes. Figure~\ref{fig:projections1} shows the result from a single discriminant subspace using all training observations and all variables to compute $\hat \Sigma_{in}^{-1}\hat\Sigma_{out}$, while Figure~\ref{fig:projections2} shows the aggregated discriminant subspace from an ensemble of 100 models with $q_0 = \lfloor 0.75 d\rfloor$ and $q = \lceil 0.5 q_0 \rceil$. In both cases $k$ was set to three. Both show fairly clear separation of the majority of classes from others, with the ensemble showing these more clearly. For comparison we have also included the discriminant projections arising from LDA and MDA in Figure~\ref{fig:projections3} and Figure~\ref{fig:projections4}, respectively. LDA shows good separation of the segmentation data set, but less so for the digits data set, while for MDA it is the reverse.

\section{Experimental Results}\label{sec:experiments}

In this section we present the results from a very large set of experiments, documenting the strong performance of the proposed approach. 

\subsection{Data Sets}

For our experiments we considered all 162 classification data sets in the Penn Machine Learning Benchmarks database~\citep{PMLB}. We repeatedly sampled training and test sets from each data set, with training sets constituting 70\% and test sets the remaining 30\%. The number of repetitions varied by overall sample size, $n$, as follows:
\begin{itemize}
    \item $0 < n < 500:$ 50 repetitions
    \item $500 \leq n < 1000:$ 20 repetitions
    \item $1000 \leq n < 5000:$ 10 repetitions
    \item $5000 \leq n:$ 5 repetitions
\end{itemize}
The exception to the above was that, due to the very large amount of compute time required for all experiments, training sets were capped at 7000 points and test sets at 3000 points (however, a different total 10 000 points was used in each repetition, where relevant).

Before passing the data sets to the different algorithms and models for fitting and prediction, all categorical variables were first one-hot-encoded. The only exception to this was in the case of the random forest models, since decision trees are able to handle categorical variables directly.

\subsection{Classification Models and Tuning}

Below we give a brief overview of the different models used for comparison, as well as how model selection was conducted for each. Although our primary interest is in the comparison between the proposed approach and the Random Forest models, we also include a number of alternatives for additional benchmarks and context.

\begin{itemize}
    \item BOPNN: The proposed approach (Bag Of Projected Nearest Neighbours), where each ensemble comprised a bag of 100 \knn~models (i.e., $B=100$). For each data set and training sample, thirty values for $k$ (the number of neighbours); $q_0$ (the size of the random subset of covariates sampled for each model); and $q$ (the number eigenvectors of each $\hat\Delta_b$ retained) were sampled uniformly as $k \sim U(\{1, ..., 5\}); q_0 \sim U(\{\lfloor p^{1/2}\rfloor, ..., \min\{\lfloor 10p^{1/2}\rfloor, p\}\});$ and $q|q_0 \sim U(\{\lceil 0.5q_0\rceil, q_0\})$ respectively. Models were fit for each setting of these hyperparameters and with the size of each bootstrap sample being 0.63 times the size of the training set ($\pi_B = 0.63$). The model with the highest Out-Of-Bag estimate for classification accuracy was then applied on the test set.
    \item BOpNN: Equivalent to above, except no discriminant subspace was found for each model (or equivalently $q$ was always set to $q_0$). This variant is included primarily to give a clear indication of the benefit of including an adaptive learning step (the determination of a discriminant subspace) within a bagged model of otherwise lazy learners. %In addition, this model gives some indication, through its comparison with BNN$_\infty$, below, whether there is benefit in purely randomised variable ``selection'' for bagged \knn~models. Note that 
    \item BNN$_{\infty}$: A bagged 1-NN model where the proportion of the sample included in each bootstrap sample is set equal to
    \begin{align*}
        \pi_B = \left(2\Gamma\left(2 + \frac{2}{p}\right)^2\right)^{\frac{p}{p+4}}\frac{1}{\hat k},
    \end{align*}
    where $\hat k$ is an estimate of the optimal value for $k$ in a single \knn~model, based on the leave-one-out cross-validation estimate for classification accuracy. This setting is a plug-in estimate for the asymptotically optimal value~\citep{samworthBNN}. Note that for $\hat k = 1$ this proportion is greater than one, and in this case we simply set $\pi_B = 0.9$.
    \item BNN: As above except where $\pi_B$ is tuned using OOB performance. This is included for comparison with BNN$_{\infty}$, to test the appropriateness of the plug-in estimate for the asymptotically optimal value for $\pi_B$.
    \item $k$NN: A single \knn~model with $k$ selected using the leave-one-out cross-validation estimate of classification accuracy.
    \item ES$k$NN: The \knn~ensemble which combines a selection from a large number of models fit to different bootstrap samples with different random subsets of the covariates~\citep{esknn}. Unlike the bagged models above, this approach suffers unless the total number of \knn~models is very large. Moreover, since no relevant OOB estimates for performance are available, the compute time required for this approach was substantially greater than any of the other methods. As a result, we used a single 25\% validation set for estimating performance and only selected $k$ from the set $\{1, ..., 10\}$. We fixed all other parameters equal to their defaults in the {\tt ESKNN} package~\citep{CRANESKNN}, previously on the Comprehensive R Archive Network~(CRAN). These settings deviated from the approach used by~\cite{esknn} only in that a single validation set, as opposed to cross-validation, was performed to select $k$, and we used 501 initial models (as the default in the package) instead of 1001. Even with these changes, experiments with this approach required substantially larger compute time than any of the other methods. We also did not observe appreciably superior performance when using 1001 models instead of 501.
    \item RF: The random forest classifier, as implemented in the {\tt R}~\citep{R} package {\tt randomForestSRC}~\citep{rfsrc}, available on CRAN. Following the same approach as for BOPNN, hyperparameter selection was conducted from 30 random selections based on OOB performance. The parameters which were tuned are ``mtry'' (the number of randomly selected covariates selected as candidates for each split in a tree), which was selected from the interval $[0.1p^{1/2}, \min\{p, 10p^{1/2}\}]$; and the minimum size of a leaf node in a tree, from $\{1, ..., 10\}$.
    \item SVM: The Support Vector Machine classifier, where multiple classes were handled using the one-vs-one approach. We used the {\tt LiquidSVM} implementation~\citep{liquidSVM}, which uses a fast technique to approximate the kernel matrix but has nonetheless shown excellent performance in comparison with exact methods~\citep{liquidSVM}. We used the default tuning grid and cross-validation settings provided in the implementation.
    \item RDA: Regularised Discriminant Analysis~\citep{RDA}, where each class is modelled using a Gaussian distribution and class probabilities are determined using Bayes rule. The means of the component distributions, $\hat{ \boldsymbol{\mu}}_1, ..., \hat{ \boldsymbol{\mu}}_K$, are determined by the sample means of the points from each class, while the covariance matrix of the $j$-th component is set equal
    \begin{align*}
        \hat \Sigma_j(\lambda, \gamma) &:= (1-\gamma)\tilde \Sigma_j(\lambda) + \frac{\gamma}{p}\mathrm{trace}(\tilde \Sigma_j(\lambda))\mathbf{I},\\
        \tilde \Sigma_j(\lambda) :=& \frac{1-\lambda}{(1-\lambda)n_j + \lambda n}\Bigg(\sum_{i: y_i = j}(\x_i - \hat{\boldsymbol{\mu}}_j)(\x_i - \hat{\boldsymbol{\mu}}_j)^\top\\
        &+\frac{\lambda}{1-\lambda}\sum_{i=1}^n(\x_i - \hat{\boldsymbol{\mu}}_{y_i})(\x_i - \hat{\boldsymbol{\mu}}_{y_i})^\top\Bigg),
    \end{align*}
    where $\lambda$ and $\gamma$ are hyperparameters which must be chosen, and for which we used 5-fold cross-validation.
\end{itemize}

\subsection{Summarising Classification Performance}

In this subsection we provide an overview of the classification performance of all methods across the collection of 162 data sets used for comparison.
In order to combine the results from different data sets, which may have vastly different characteristics and represent classification tasks of varying difficulty, we first standardise the results. We consider two different standardisations, and apply them to the classification accuracy values from the collection of models obtained on each data set, and each repetition of the sampling of training/test splits. For each model and each data set, we then take the two averages of its standardised accuracy values across the different test sets as the model's performance measures for that data set. Specifically, if $A_{m,i,t}$ represents classification accuracy of the $m$-th model on the $t$-th train/test split from the $i$-th data set, then we compute
\begin{align}
    A^*_{m,i,t} &:= \frac{A_{m,i,t} - \min_o{A_{o,i,t}}}{\max_o{A_{o,i,t}}-\min_o{A_{o,i,t}}}\\
    A^{**}_{m,i,t} &:= \frac{A_{m,i,t} - \overline{A_{\cdot,i,t}}}{s(A_{\cdot,i,t})},
\end{align}
where $\overline{A_{\cdot,i,t}}$ and $s(A_{\cdot,i,t})$ are the average and standard deviation of the accuracy values from all methods on the $t$-th train/test split of data set $i$. In the case of $A^*_{m,i,t}$ this simply maps the accuracy values to the interval $[0,1]$, while $A^{**}_{m,i,t}$ is the common studentised value for $A_{m,i,t}$. The performance values for each method on a given data set are then just the averages of these standardised accuracies over $t$.

%\end{multicols}

\begin{figure*}
    \centering
    \subfigure[{Accuracy values mapped to the [0,1] interval}]{\includegraphics[width=.48\linewidth]{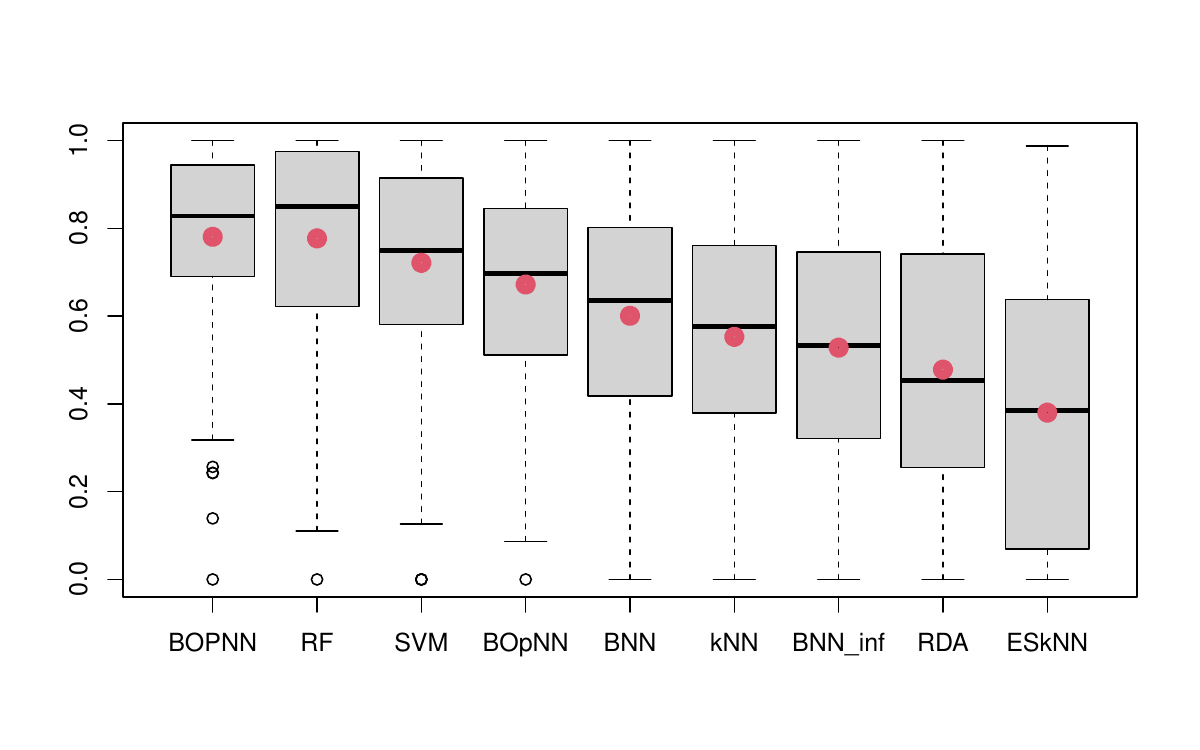}}
    \subfigure[Studentised accuracy values]{\includegraphics[width=.48\linewidth]{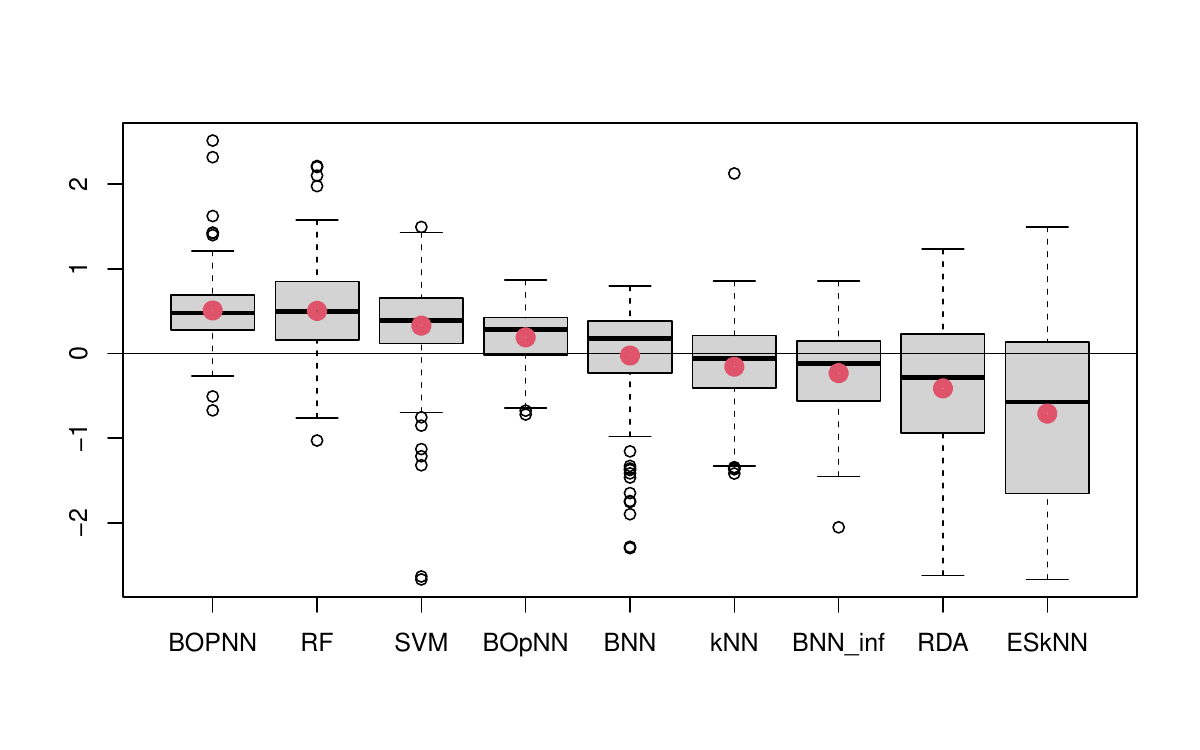}}
    \caption{Boxplots of accuracy distributions for different classification models, using two different standardisations.}
    \label{fig:boxplots}
\end{figure*}

%\begin{multicols}{2}

\subsubsection{Accuracy Distributions}

The distributions of standardised accuracy measures across all 162 data sets are shown in Figure~\ref{fig:boxplots}. The main take-aways are summarised as
\begin{enumerate}
    \item The averages of standardised accuracy values of BOPNN and RF are extremely similar, with BOPNN having slightly higher average but with RF having wider distributions. This is noteworthy since RF classifiers have commonly been referred to as excellent general purpose models; seldom substantially worse than any alternatives. These results suggest BOPNN similarly enjoys this feature, with arguably a better ``worst-case'' than RF due to a similar average and narrower distribution.
    \item SVM outperforms all methods except the bagged models of adaptive non-parametric smoothers (BOPNN and RF).
    \item BOpNN is substantially inferior to BOPNN, showing the importance of the adaptive learning step within the bagged model.
    \item BOpNN is substantially superior to BNN. Although these two bagged models were tuned over disjoint collections of hyperparameters (BOpNN over $q_0$ and $k$, and BNN over $\pi_B$) the magnitude of the difference in performance is some indication that the purely randomised variable ``selection'' does indeed offer an improvement over its exclusion. Additional experiments are warranted, but as this is peripheral to our investigation and the number of experiments and results being presented is already substantial, we do not pursue this further here.
    \item BNN is substantially superior to BNN$_\infty$. This suggests either that it is inappropriate to rely on the asymptotic theory for relatively small samples, or that selection of $\hat k$ is subject to too much variation to be a reliable plug-in value.
    \item ES$k$NN performed very poorly, and was the worst performing model on a large proportion of the data sets. It is worth noting that the code released by the authors (previously on CRAN) included minor errors, such as populating arrays as though they were lists. we took care to correct these appropriately, however it is not impossible that errors were made. Having said this, the approach was the best performing method in some cases, which is arguably unlikely if an inappropriate correction of the code was made.
\end{enumerate}

%\end{multicols}

\begin{figure*}
    \centering
    \subfigure[{Accuracy values mapped to the $[0,1]$ interval}]{\includegraphics[width=0.35\linewidth]{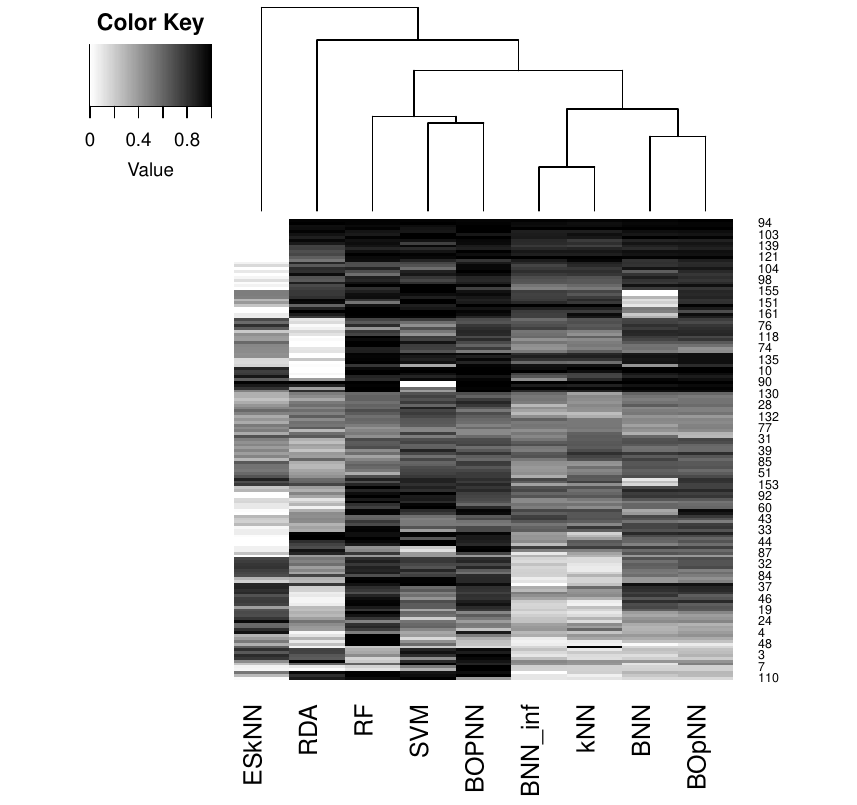}}
        \subfigure[Studentised accuracy values]{\includegraphics[width=0.35\linewidth]{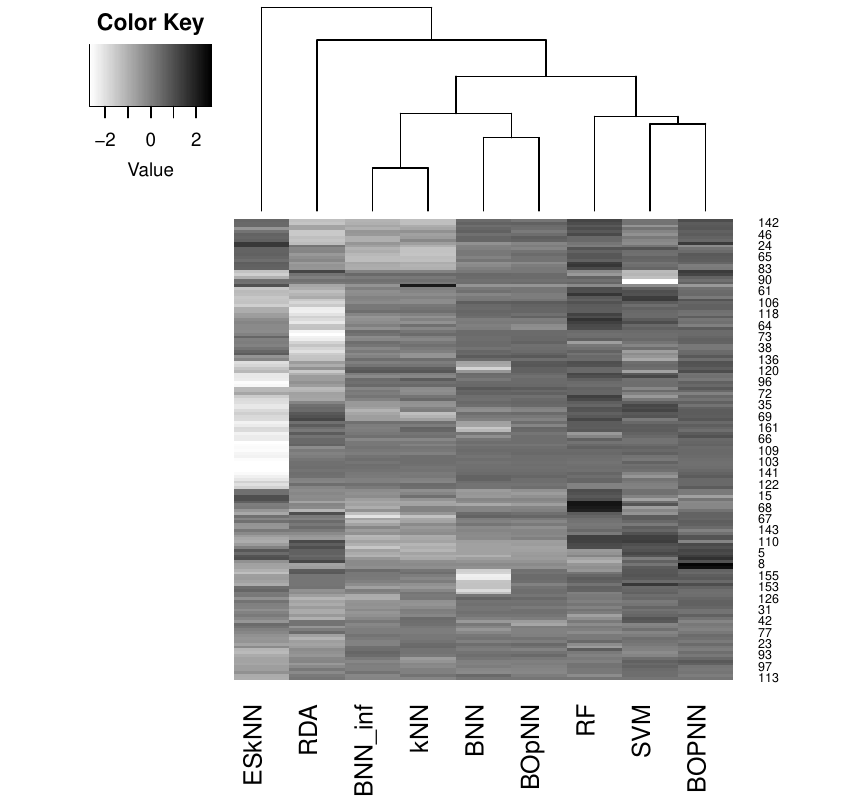}}
    \caption{Standardised classification accuracy across all data sets}
    \label{fig:all_accuracy}
\end{figure*}

%\begin{multicols}{2}

\subsubsection{Pairwise Comparisons}

Although the accuracy distributions provide a strong indication of overall performance of the different methods, they do not directly show the comparative performance of any two methods on any specific data sets. Figure~\ref{fig:all_accuracy} shows the standardised accuracy values of all methods across all data sets. What is interesting is that the dendrogram along the top axis of each sub-figure shows the performance of BOPNN is mist similar to that of SVM. Naturally the overall magnitude of the performances will tend to lead to high similarity between the overall performances of the different methods, however the specific data sets on which BOPNN and SVM perform better/worse overcomes the fact that BOPNN and RF have appreciably more similar average performance than does either of them to SVM.\\
\\
In addition Table~\ref{tb:pairwise} shows the number of times, out of the total 162, the method listed row-wise achieved significantly superior performance to the method listed column-wise on a given data set. Significance was determined based on a paired Wilcoxon signed rank test~\citep{wilcoxon1992individual} with test size 0.05\footnote{We acknowledge the arbitrariness of this test size, and do not mean to indicate any statistical relevance of these comparisons. Rather, we mean only to give a sense of the frequency with which each method outperforms each other method, while appropriately accounting for some of the randomness inherent in such a comparison.}.

\begin{table*}[]
    \centering
    {\smaller
    \begin{tabular}{r|p{1.1cm}p{1.1cm}p{1.1cm}p{1.1cm}p{1.1cm}p{1.1cm}p{1.1cm}p{1.1cm}p{1.1cm}}
         &BOPNN & RF & SVM & BOpNN & BNN & \knn & BNN$_\infty$ & RDA & ES$k$NN\\
         \hline
BOPNN & 0 & 37 & 53 & 60 & 70 & 99 & 107 & 115 & 107\\
RF & 46 & 0 & 57 & 71 & 84 & 101 & 101 & 108 & 114\\
SVM & 33 & 35 & 0 & 62 & 76 & 94 & 98 & 105 & 100\\
BOpNN & 6 & 19 & 33 & 0 & 23 & 73 & 76 & 92 & 92\\
BNN & 4 & 13 & 26 & 1 & 0 & 66 & 72 & 84 & 84\\
\knn & 6 & 19 & 21 & 7 & 25 & 0 & 16 & 72 & 81\\
BNN$_\infty$ & 3 & 15 & 17 & 3 & 20 & 8 & 0 & 64 & 85\\
RDA & 9 & 18 & 15 & 29 & 41 & 44 & 45 & 0 & 64\\
ES$k$NN & 8 & 11 & 17 & 21 & 31 & 41 & 45 & 55 & 0\\

    \end{tabular}}
    \caption{Pairwise comparative accuracy. Values in table indicate the number of times the method listed row-wise significantly outperformed the method listed column-wise. For example, BOPNN significantly outperformed RF 37 times, while RF significantly outperformed BOPNN 46 times. Significance was determined using a paired Wilcoxon signed rank test, with size 0.05.}
    \label{tb:pairwise}
\end{table*}

Once again we are particularly interested in the comparison between BOPNN and Random Forests. Although the average performance of BOPNN is slightly superior, as shown in the previous subsection, we see here that RF outperformed BOPNN more frequently than the reverse. What is interesting to note is that RF both outperforms the majority of the other methods more often than does BOPNN, and is also more frequently outperformed \textit{by} them; indicated by greater values in the RF row \textit{and} column than those of BOPNN.

\subsection{Relationships between Performance and ``Meta-data''}

Similar to~\cite{hofmeyr2024optimal} we investigate the relationships between the relative performance of the different methods, and the characteristics of the data. That is, each data set is characterised by
\begin{enumerate}
    \item $n$: the number of observations
    \item $d$: the total number of variables after one-hot encoding
    \item cat\_rat: the proportion of binary variables in the one-hot encoded data
    \item $K$: the number of classes
    \item imbal: the class imbalance, defined as the variance of the class proportions
    \item compl: a measure of the complexity of the class decision boundaries, defined as $\log\left(\frac{A_{1NN}}{A_{NC}}\right)$, where $A_{1NN}$ is the leave-one-out cross-validation (LOOCV) estimate for the accuracy of the 1-nearest-neighbour classifier on the data, and $A_{NC}$ is the LOOCV estimate for the accuracy of the nearest centroid classifier\footnote{The nearest centroid classifier simply classifies a point to the class whose mean vector is closest.},
\end{enumerate}
\noindent
and we compute the marginal correlations between the studentised accuracy of each method over all data sets and the data set characteristics, as well as the Ordinary Least Squares (OLS) linear regression coefficients after standardising all variables\footnote{we used studentised performance instead of the $[0, 1]$ mapped performance as their distributions are closer to Gaussian and, all other things being equal, may therefore be more appropriate when used in quantifying linear relationships}. These OLS coefficients give an indication of the correlations between the data set characteristics and the studentised performance \textit{after accounting for the values of the other data set characteristics}.

\begin{figure*}
    \centering
    \includegraphics[width=.6\linewidth]{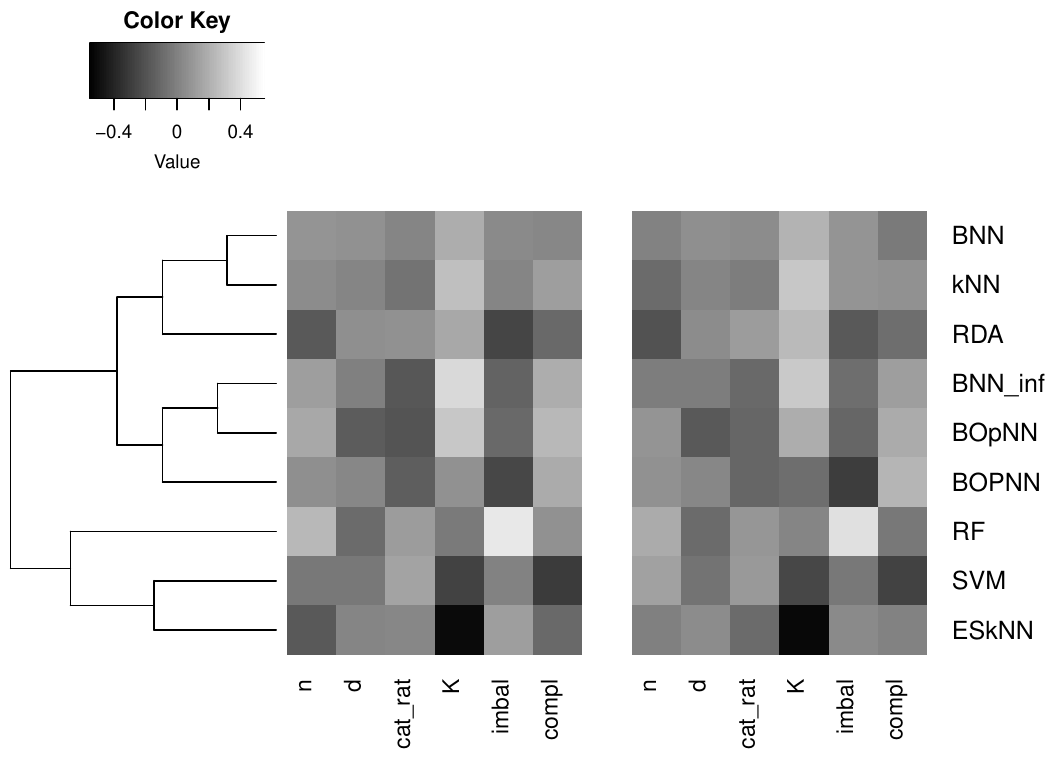}
    \caption{Correlations between studentised accuracy and data set characteristics. Left: marginal correlations, Right: OLS coefficients}
    \label{fig:heatmaps}
\end{figure*}

Figure~\ref{fig:heatmaps} shows heatmaps indicating the strength of these correlations, with the marginal correlations in the left heatmap and the OLS coefficients in the right. Because these relationships are determined between the data set characteristics and a \textit{standardised} accuracy measure, they give an indication of the relationships between the different methods and data set characteristics \textit{relative to the other methods considered}. For example, as noted by~\cite{hofmeyr2024optimal}, we expect all methods will perform relatively better on larger data sets, all other things being equal. This would correspond with light colours in the first column of the right heatmap (positive OLS coefficients), however some methods are better/worse at leveraging larger samples than others and this is reflected by both positive AND negative OLS coefficients in the relationships with standardised accuracy. The lightest colours in this column indicate that RF, SVM and BOpNN may be relatively better at leveraging larger samples than the other methods. Unsurprisingly the parametric RDA apparently leverages larger samples the least, indicated by the darker colour in this column. Some other noteworthy take-aways, for the purpose of this investigation, are:
\begin{enumerate}
    \item BOPNN has its strongest negative relationship with class imbalance, indicating that the proposed approach is less effective than other methods at handling this class imbalance. It is also noteworthy that RF performs particularly well on imbalanced cases. The adaptive component of BOPNN comprises a linear transformation through projections into discriminant spaces, and so applies globally. The local, non-linear component of BOPNN comes subsequently from the application of \knn~on the transformed data. When the classes are especially imbalanced it is likely that weighing the emphasis of the classes on the matrices $\hat \Sigma_{in}$ and $\hat \Sigma_{out}$ equally, i.e., by using instead the formulations $\hat \Sigma_{in} = \frac{1}{K}\sum_{k=1}^K \frac{1}{n_k}\sum_{i:y_i = k} (\x_i - \x_{i_k})(\x_i - \x_{i_k})^\top$ and $\hat \Sigma_{out} = \frac{1}{K}\sum_{k=1}^K \frac{1}{n_k}\sum_{i:y_i = k} (\x_i - \x_{i'_k})(\x_i - \x_{i'_k})^\top$, respectively, where $n_k$ is the number of observations in class $k$, may improve performance in such cases.
    \item BOPNN has its strongest positive relationship with the ``complexity'' of the decision boundaries. It is possible that this observation is somewhat artificial, given that the meausre of complexity is governed by the performance of the 1NN model. However, this performance is quantified relative to the other methods being compared, which includes numerous other nearest neighbour based methods.
    \item Compared with Random Forests, BOPNN may be better suited to higher dimensional examples but may leverage larger samples less well.
    \item The simplistic one-hot-encoding followed by Euclidean distance calculation currently employed in BOPNN may be inappropriate, as indicated by its comparatively poor performance when a large number of categorical variables are present. Although alternative distance metrics, which combine numeric and categorical covariates better, are available, we prioritised computational speed and only the Euclidean metric allows for very fast nearest neighbour search.
\end{enumerate}

%\end{multicols}

%\begin{multicols}{2}

\section{Concusions}\label{sec:conclusions}

In this paper we discussed the importance of including an adaptive learning step within the context of bootstrap aggregating, or ``bagging'', and proposed a simple adaptive \knn~model, in which neighbours are determined within a discriminant subspace designed to enhance the separation of classes as captured by \knn, for use within ``bagged'' ensembles. The discriminant subspace framework naturally leads to measures of variable importance and offers instructive visualisation of the classes through projections into the discriminant subspaces (or an aggregated variant incorporating the entire ensemble).

In an extensive set of experiments we documented the strong potential offered by the proposed approach. Noteworthy findings are that across varied contexts the proposed approach is more or less on par with Random Forests, on average, but that the particular data sets on which the proposed approach may be more or less suited in fact align better with Support Vector Machines. Potential directions for improvement of BOPNN include an alternative, but computationally efficient, way to incorporate categorical variables, as well as strategies to enhance performance with highly imbalanced class proportions.

\bibliographystyle{ieeetr}

%\bibliography{Bibliography-MM-MC}


%\begin{IEEEbiographynophoto}{David P. Hofmeyr}
%Received masters degrees from Edinburgh and Lancaster Universities, and a PhD from Lancaster University. He is currently with the School of Mathematical Sciences at Lancaster University, and was previously with the department of statistics and actuarial science at Stellenbosch University, South Africa. His research interests are primarily in the areas of multivariate and non-parametric statistics.



%\vfill


\end{multicols}
\end{document}